\documentclass[10pt,twocolumn,letterpaper]{article}

\usepackage{amsmath,amsfonts,bm}

\def\eqref#1{equation~\ref{#1}}

\def\1{\bm{1}}

\def\ve{{\bm{e}}}

\def\vh{{\bm{h}}}

\def\vs{{\bm{s}}}

\def\vx{{\bm{x}}}

\def\vz{{\bm{z}}}

\def\mW{{\bm{W}}}

\DeclareMathAlphabet{\mathsfit}{\encodingdefault}{\sfdefault}{m}{sl}
\SetMathAlphabet{\mathsfit}{bold}{\encodingdefault}{\sfdefault}{bx}{n}

\def\gL{{\mathcal{L}}}
\def\gM{{\mathcal{M}}}

\def\gR{{\mathcal{R}}}

\def\gV{{\mathcal{V}}}

\newcommand{\R}{\mathbb{R}}

\DeclareMathOperator*{\argmax}{arg\,max}

\usepackage[pagenumbers]{cvpr} 

\usepackage{graphicx}
\usepackage{amsmath}
\usepackage{amssymb}
\usepackage{booktabs}
\usepackage{multirow}
\usepackage{enumitem}

\usepackage{xcolor}
\usepackage[pagebackref,breaklinks,colorlinks]{hyperref}

\usepackage[capitalize]{cleveref}
\crefname{section}{Sec.}{Secs.}
\Crefname{section}{Section}{Sections}
\Crefname{table}{Table}{Tables}
\crefname{table}{Tab.}{Tabs.}

\definecolor{deemph}{gray}{0.6}
\newcommand{\gc}[1]{\textcolor{deemph}{#1}}
\definecolor{teachercolor}{gray}{.9}

\usepackage{pifont}%
\newcommand{\cmark}{\ding{51}\xspace}%
\newcommand{\xmark}{\ding{55}\xspace}%

\def\alambic{\includegraphics[width=0.03\linewidth]{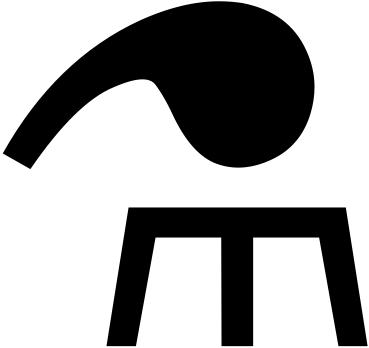}\xspace}
\def\alambictwocol{\includegraphics[width=0.015\linewidth]{./fig/alembic-crop.pdf}\xspace}

\newcommand\our{G2SD}

\newcommand\deitiii{DeiT \uppercase\expandafter{\romannumeral3}}

\newcommand{\ourgd}{G2SD w/o S.D}
\newcommand*\samethanks[1][\value{footnote}]{\footnotemark[#1]}

\def\cvprPaperID{4881} 
\def\confName{CVPR}
\def\confYear{2023}

\begin{document}

\title{Generic-to-Specific Distillation of Masked Autoencoders}
\author{%
{Wei Huang$^{1,}$\thanks{~Equal contribution. $\S$ Contribution during internship at Microsoft Research. $\dagger$ Corresponding authors.},~~Zhiliang Peng$^{1, \S,}$\samethanks[1],~~Li Dong$^{2}$,~~Furu Wei$^{2}$,~~Jianbin Jiao$^{1, \dagger}$,~~Qixiang Ye$^{1, \dagger}$
} \\
University of Chinese Academy of Sciences$^{1}$ \\
Microsoft Research$^{2}$ 
}
\maketitle

\begin{abstract}
Large vision Transformers (ViTs) driven by self-supervised pre-training mechanisms  achieved unprecedented progress.
Lightweight ViT models limited by the model capacity, however, benefit little from those pre-training mechanisms.
Knowledge distillation defines a paradigm to transfer representations from large (teacher) models to small (student) ones. However, the conventional single-stage distillation easily gets stuck on task-specific transfer, failing to retain the task-agnostic knowledge crucial for model generalization.
In this study, we propose generic-to-specific distillation (G2SD), to tap the potential of small ViT models under the supervision of large models pre-trained by masked autoencoders.
%
In generic distillation, decoder of the small model is encouraged to align feature predictions with hidden representations of the large model, so that task-agnostic knowledge can be transferred. 
In specific distillation, predictions of the small model are constrained to be consistent with those of the large model, to transfer task-specific features which guarantee task performance.
With G2SD, the vanilla ViT-Small model respectively achieves 98.7\%, 98.1\% and 99.3\% the performance of its teacher (ViT-Base) for image classification, object detection, and semantic segmentation, setting a solid baseline for two-stage vision distillation. 
Code will be available at \url{https://github.com/pengzhiliang/G2SD}.
\end{abstract}

\section{Introduction}
\label{sec:intro}

Vision transformers (ViTs)~\cite{vit,scalingViT} have been promising representation models, particularly when trained upon large-scale datasets using self-supervised learning methods~\cite{dino}. The masked image modeling (MIM) methods~\cite{mae, beit}, which train representation models by reconstructing pixels~\cite{mae,simmim,HiViT}, tokens~\cite{beit,beitv2,cae} or features~\cite{maskfeat,data2vec}, promoted the performance of large ViT models to a new height. 

However, when acclaiming the promising performance of large ViT models, we notice that small ViT models, $e.g.$, ViT-Tiny and ViT-Small, unfortunately, benefit little from either the big training data or self-supervised learning methods. For example, the ViT-Large model trained by MAE~\cite{mae} outperforms the CNN model~\cite{convnext} by 1.6 points on ImageNet-1k, while the ViT-Small model is inferior to its CNN counterpart~\cite{convnext}.
%
In most scenarios with limited computational resources, $e.g.$, front-end recognition systems, CNNs~\cite{resnet,MobileNetV3} remain the preferred models.

\begin{figure}[t]
\centering
\includegraphics[width=1\linewidth]{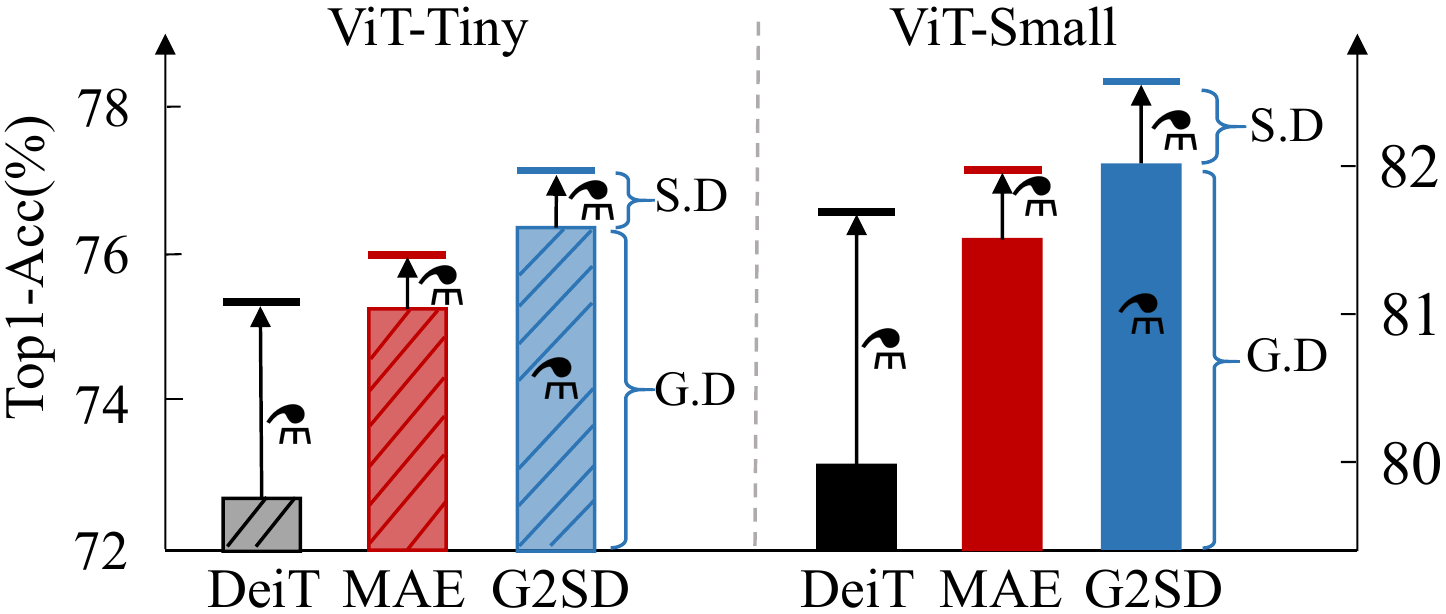}
\caption{
Comparison of single-stage distillation models (from scratch~\cite{deit} and pre-trained by the self-supervised method MAE~\cite{mae}) with the two-stage distillation counterparts (G2SD) using the same teacher model. G.D and S.D respectively denote generic and specific distillation. 
\alambic is the symbol of distillation. }
\label{fig:figure-1}
\end{figure}

\textit{Do vanilla small ViT models really have no future?}
%
We attempt to answer this question from the perspective of knowledge distillation in this study. To fulfill this purpose, the first step is to revisit the conventional knowledge distillation methods~\cite{hinton2015distilling,fitnet,deit} in the age of supervised learning. It is observed that task-oriented distillation~\cite{deit} reports unsatisfactory performance,~\cref{fig:figure-1}. One reason could be that this kind of task-oriented distillation only focus on task-specific knowledge while missing some kind of task-agnostic knowledge which is beneficial to generalization ability improvement and can be effectively endowed by self-supervised teacher model.
In natural language processing, two-stage distillation method, $e.g.$, TinyBERT~\cite{tinybert}, was exploited to overcome the limitation and transfer generic knowledge embedded from teacher to student models. Nevertheless, whether or not this paradigm applicable to vision tasks remains unexplored. 

\begin{figure*}[t]
\centering
\includegraphics[width=1\linewidth]{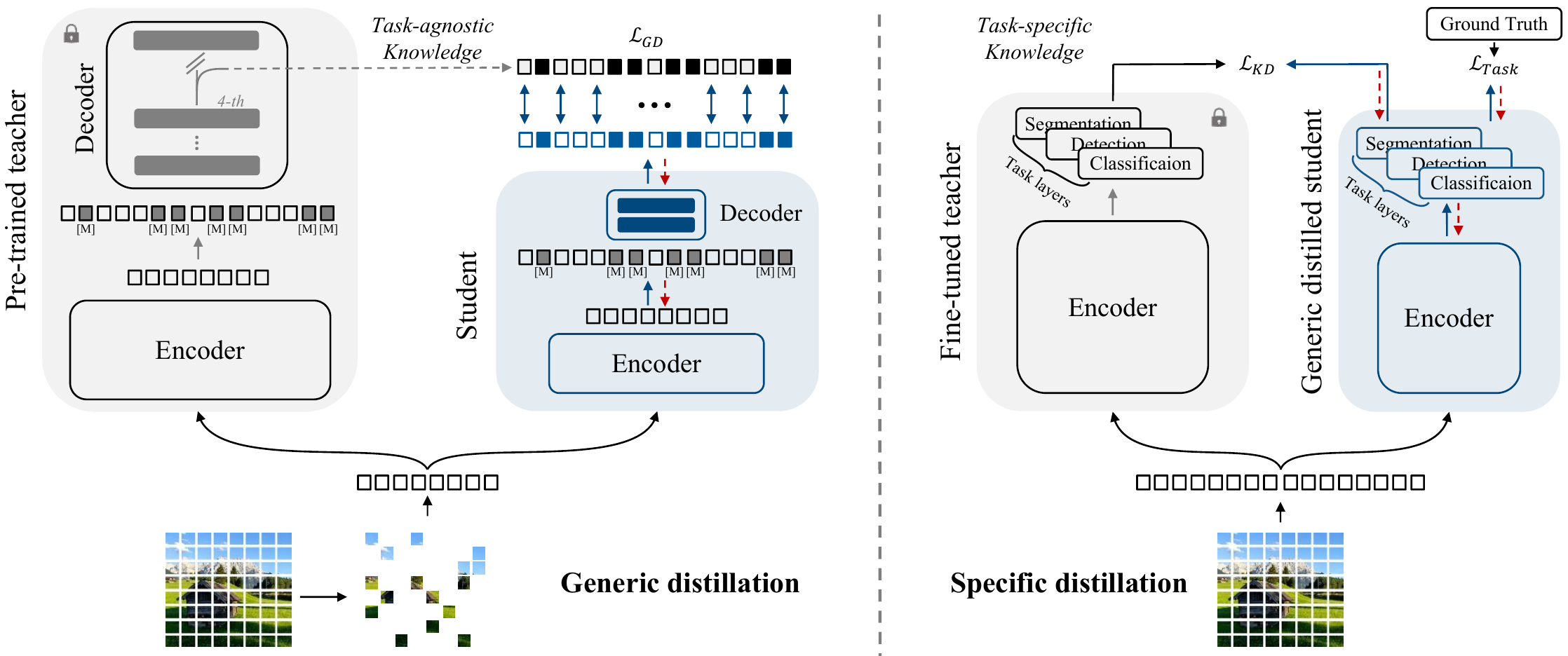}
\caption{Diagram of the proposed generic-to-specific distillation (G2SD). [M] denotes mask token. In the generic distillation stage (\textit{left}), masked images are converted to patches and fed to both the teacher and student encoders for feature extraction. Feature predictions of the student decoder are aligned with those of the teacher at both visible and predicted patches. In the specific distillation stage (\textit{right}), student models are trained to have consistent predictions with teacher models fine-tuned on the specific task.}
\label{fig:framework}
\vspace{-1em}
\end{figure*}

In this study, we aim to establish a general-to-specific distillation baseline for vision tasks based on sophisticated self-supervised learning ($e.g.$, MAE~\cite{mae}), to guarantee that lightweight ViTs can simultaneously soak up task-agnostic and task-specific representations from teacher models for greater generalization and higher task performance, Fig.~\ref{fig:figure-1}. 
Specifically, at the generic distillation stage, a student model is encouraged to obtain the task-agnostic knowledge from the teacher models. 
%
The encoder and decoder of pre-trained MAE constitute the teacher model while a light-weight decoder is attached to the lightweight vision Transformer as the student model, ~\cref{fig:framework}. The input image is randomly partitioned to visible and masked patches. The visible patches are fed to encoders. The hidden feature outputs of teacher decoder's intermediate layer is used to guide training of the student model.
%
For task-specific distillation, the fine-tuned MAE model equipped with task layers~\cite{deit,maskrcnn, upernet} teaches student model the task-specific knowledge (\eg, classification score). The student backbone is initialized from the previous distillation stage while the task layers are randomly initialized. Predictions of the student are constrained to be consistent with those of the teacher as well as ground truth labels. Such a task-specific distillation phase guarantees the performance of downstream tasks, \eg, image classification, object detection and semantic segmentation. 
%

With G2SD, the vanilla ViT-Small model with \textbf{~26\%} parameters and \textbf{2.6$\times$} throughput of the ViT-Base teacher, obtains 1) \textbf{98.6\%} (82.5\% \vs 83.6\%) top-1 accuracy of its teacher on ImageNet-1k~\cite{imagenet} for image classification task, 2) \textbf{98.1\%} (50.6 \vs 51.6) mAP of its teacher on MS COCO~\cite{mscoco} for object detection and 3) \textbf{99.3\%} (48.0 \vs 48.3) mIoU of its teacher on ADE20k~\cite{ade20k} for semantic segmentation.
Furthermore, G2SD demonstrates better generalization ability than its single-stage distillation counterparts in terms of occlusion invariance and robustness.

%
The contributions are summarized as follows:
\begin{itemize}[leftmargin=1.5em]
\item We propose general-to-specific distillation (\our{}) to transfer task-agnostic and task-specific knowledge from masked autoencoders to lightweight ViTs, setting a solid baseline for two-stage vision model distillation.

\item We design a simple-yet-effective generic distillation strategy by aligning the student's predictions with hidden features of the pre-trained masked autoencoder at visible and masked patches.

\item Experiments show that the lightweight student model with \our{} achieves competitive results across vision tasks, improving the performance of lightweight ViT models to a new height.

\end{itemize} 

\section{Related Work}
\label{sec:related:work}
\paragraph{Vision Transformers.}

ViTs~\cite{vit} have achieved impressive performance across vision tasks~\cite{scalingViT,conformer,beit,mae,beitv2,imted,vitdet}. Furthermore, ViTs demonstrated the superiority in terms of robustness and generalization~\cite{naseer2021intriguing,mae,beitv2}, compared to their CNN counterparts.  However, due to the lack of inductive bias, ViTs report unsatisfactory performance in the limited model capacity regime~\cite{vit,deit}. One solution is to explicitly introduce convolutional operators to ViTs~\cite{MobileViT,TinyViT} to enhance the competitiveness compared to lightweight CNNs~\cite{MobileNetV3}.
The other way is using large models act as teachers to transfer inductive bias to ViTs in the knowledge distillation fashion~\cite{deit,Chen2022DearKDDE,TinyViT}. This study focuses on the latter.

\paragraph{Self-supervised Learning.}
To explore big data without high-quality labels, self-supervised learning has been the preferred paradigm to construct representation models~\cite{dino}. Masked language modeling~\cite{bert} achieved great success in natural language process (NLP) field. Inspired by it, BEiT~\cite{beit} introduced the \textit{mask-then-predict} paradigm to the computer vision filed and exploited the great potential of masked image modeling (MIM) on various tasks. BEiT v2~\cite{beitv2} constructed a semantic-rich visual tokenizer in order to get better target. MAE~\cite{mae} set a new baseline for MIM by reconstructing pixels at mask patches with a decoupling encoder-decoder architecture. Meanwhile, feature masking and reconstruction methods~\cite{maskfeat,data2vec} demonstrated advantages over the pixel-reconstruction approach. Those methods, however, when exploring performance upper bound by finding better supervisions to pre-train large ViTs, ignored the adaptability of lightweight models with limited capacity. In this study, \our{} develops a two-stage knowledge distillation baseline for lightweight ViT models to enjoy MIM advantages.
%

\paragraph{Knowledge Distillation.} 
The pioneering work~\cite{hinton2015distilling} compressed the ``dark knowledge" from a large (teacher) model to a small (student) model by minimizing KL divergence between the output logits distribution of the two models. NKD~\cite{nkd} rethinked the relation between knowledge distillation loss and the original cross-entropy and proposed a new KD loss. FitNet~\cite{fitnet} pioneered feature distillation by utilizing the intermediate layers' features from the teacher model. To find the better feature layers for distillation, subsequent works~\cite{KR,L2T} studied the factor of connection path cross multiple i ntermediate layers between teacher and student networks. Besides distilling the knowledge contained in samples, inter-samples relation, as structural information, was transferred to student models~\cite{SP,RKD,CCKD}. Knowledge distillation has also been elaborately studied for ViTs~\cite{deit,Yang2022ViTKDPG,Chen2022DearKDDE,manifold_distill,MetaKD}.
%
SSTA~\cite{SSTA} simultaneously learned from the supervised teacher and self-supervised teacher, which was regarded as teaching assistant. 
However, those methods are designed and evaluated on specific task, such as classification, detection~\cite{frs} or segmentation~\cite{cwd}.
The task-oriented methods experience difficulty in transferring task-agnostic knowledge, while task-agnostic knowledge is crucial to guarantee the generalization ability of lightweight models. To overcome the limitation, TinyBERT~\cite{tinybert} pioneered two-stage knowledge distillation in natural language processing. Nevertheless, the problem remains to be explored in vision tasks. 
%
In this study, we focus on excavating task-agnostic knowledge embedded in masked autoencoders to establish a solid baseline for vision model distillation in the era of self-supervised learning.


\section{Preliminary}
\label{sec.pre}

\paragraph{Transformer Representations.} 
To learn visual representations, ViT~\cite{vit} converts each image to a sequence of `words' (vectors) by partitioning it to patch grid.
In specific, the input image $\vx \in \R^{H \times W \times C}$ is divided to $N=(H*W)/P^2$ non-overlapping patches $\{\vx^{p}_{i} \}_{i=1}^{N}$, where $H$, $W$, $C$ and $P$ respectively denote the image height, width, channel and patch stride, and $\vx^{p}_{i} \in \R^{N \times (P^2C)}$.
In this study, a $224\times 224\times 3$ size image is reshaped to a $14\times 14$ grid of image patches, each patch size is $16\times 16\times 3$.
Meanwhile, positional information\footnote{The positional embeddings are omitted for simplicity} is embedded to the patches.
By passing the vectors through stacked Transformer blocks, which consist of a multi-head self-attention~\cite{vaswani2017attention} layer and a fully connected feed-forward network, the input vectors are converted to image representations. 

\paragraph{Masked Autoencoders.} 
\label{sec:revist_mae}
The MAE model contains an encoder $f_e$ and a decoder $f_d$, where both $f_e$ and $f_d$ are stacked Transformer blocks. 
The input tokens $\{\vx^{p}_{i} \}_{i=1}^{N}$ are grouped to the visible token set $\{\vx^{p}_{i} \}_{i\in \gV}$ and the masked token set $\{\vx^{p}_{i} \}_{i\in \gM}$.
While the visible tokens are fed to the encoder $f_e$ to extract features, the masked tokens act as the learning targets, which are required to be reconstructed during self-supervised learning (MIM).
In the MAE method~\cite{mae}, a high mask ratio (\eg, 75\%) is adopted, to prevent information leakage (\ie, simply extrapolating masked pixels from the neighbors) in the pre-training phase.

Specifically, $\{\vx^{p}_{i} \}_{i\in \gV}$ are fed to $f_e$ to obtain latent features $\{\ve_{i} \}_{i\in \gV}$, where $\ve_{i} = f_e(\vx^{p}_{i})$ for each $i\in \gV$. 
A shared learnable mask token $\ve_{[\text{M}]}$ is considered as the placeholder of tokens in $\gM$.
After that, we have the input tokens $\{\vh_{i} \}_{i=1}^{N}$ for the decoder $f_d$, where 
\begin{align}
h_i = \ve_{[\text{M}]} \odot \delta(i\in \gM) + \ve_{i} \odot (1 - \delta(i\in \gM)),
\label{eq:mae_mask}
\end{align}
and $\delta(\cdot)$ denotes an indicator function.
$\{\vh_{i} \}_{i=1}^{N}$ are then fed to $f_d$ to generate predictions at all positions $\{\vz_{i} \}_{i=1}^{N}$.
The loss is calculated by comparing the normalized pixels with predictions at masked positions $\gM$, as 
\begin{align}
\gL_{\rm{MAE}} = \sum_{i \in \gM}||\text{LN}(\vx^{p}_{i}) - \vz_{i}||_{2},
\label{eq:mae_objective}
\end{align}
where $\text{LN}(\cdot)$ is the layer normalization without affine transformation, $a.k.a$, the per-patch normalization in MAE. 
After pre-training, the encoder acts as backbone to extract representations for various tasks and the decoder is abandoned. 
As the model does not access any label in the pre-training stage, it is assumed that the features extracted by encoder are general to downstream tasks.

\section{Generic-to-Specific Distillation}
\label{sec:methods}
Our generic-to-specific distillation (G2SD) emphasizes transferring the task-agnostic knowledge embedded in large pre-trained masked autoencoders~\cite{mae}. In conjunction with task-specific distillation, G2SD endows lightweight models favorable generalization ability and competitive results.
    
\subsection{Generic Distillation: Task-agnostic knowledge Transfer}
\label{sec:general}
In each training iteration, the generic distillation consists of a feed-forward procedure of the teacher model, a feed-forward and a back-propagation procedure of the student model,~\cref{fig:framework} (\textit{left}). 
In the feed-forward procedure, outputs from an intermediate layer of the teacher decoder and the final layer of the student decoder are compared to calculate the generic distillation loss.
%

Denote the encoder and decoder of teacher model pre-trained with MAE method as $f^t_e$ and $f^t_d$, and the encoder and decoder of student models as $f^s_e$ and $f^s_d$, respectively.
Input tokens $\{\vx^{p}_{i} \}_{i=1}^{N}$ are randomly categorized to visible ones $\{\vx^{p}_{i} \}_{i\in \gV}$ and masked ones $\{\vx^{p}_{i} \}_{i\in \gM}$.
The visible tokens $\{\vx^{p}_{i} \}_{i\in \gV}$ are simultaneously fed to $f^t_e$ and $f^s_e$ to extract features $\{\ve^{t}_{i} \}_{i\in \gV}$ and $\{\ve^{s}_{i} \}_{i\in \gV}$. 
According to \cref{eq:mae_mask}, we have the input tokens set $\{\vh^{t}_{i} \}_{i=1}^{N}$ for the teacher decoder and $\{\vh^{s}_{i} \}_{i=1}^{N}$ for the student decoder.
In general, a flexible decoder consists of multiple Transformer blocks.
We respectively mark the depth of teacher decoder and that of student decoder as $L$ and $l$, where $l\leq L$ in our experiments.
Let features output by the $l$-th layer of the teacher decoder $f^t_d$ as
$\{\hat\vz^{t}_{i} \}_{i=1}^{N}$, where $\hat\vz^{t}_{i}=f^t_{d_{l}}(\vh^{t}_{i})$.
The student decoder $f^s_d$ employs $l$ Transformer blocks on $\{\vh^{s}_{i} \}_{i=1}^{N}$ and calculates the output features as $\{f^s_{d}(\vh^{s}_{i}) \}_{i=1}^{N}$.
%
Subsequently, a linear layer $\mW$ is applied on $ f^s_{d}(\vh^{s}_{i}) $ to align with the channel dimension of $\hat\vz^{t}_{i}$ and generates predictions $\vz^{s}_{i}$, \ie, $\vz^{s}_{i} = \mW f^s_{d}(\vh^{s}_{i}) $.

According to the above definitions, a generic distillation loss is defined as
\begin{align}
\gL_{\rm{GD}} = \sum_{i \in \{\gV\bigcup\gM\}}\text{Smooth-}\ell_1(\text{LN}(\hat\vz^{t}_{i}) - \vz^{s}_{i}),
\label{eq:gd_objective}
\end{align}
where $\text{Smooth-}\ell_1(\cdot)$ is a trade-off function between $\ell_1$ and $\ell_2$.  
By minimizing $\gL_{\rm{GD}}$ on the visible tokens $\gV$, the student encoder is optimized to extract features in the way like the teacher encoder, $i.e.$, mimicking feature extraction behavior.
By minimizing $\gL_{\rm{GD}}$ on the masked tokens $\gM$, the student encoder and decoder are optimized to learn context modeling ability from teacher models.
Optimizing $\gL_{\rm{GD}}$ on all tokens transfers task-agnostic knowledge.

\subsection{Specific Distillation: Task-specific Representation Configuration}
\label{sec:specific}
After generic distillation, lightweight models are able to generalize to downstream tasks and reach competitive performance, which has been validated by comprehensive experiments (See~\cref{tab:abla:two_stage}).
Nevertheless, limited by a relatively small model size and number of parameters, lightweight models still have a performance gap with their teachers. 
To bridge the gap, specific distillation is performed so that compact yet discriminative features can be configured for downstream tasks, such as image classification, object detection, and semantic segmentation.

For specific distillation, the teacher model $f^t$ is first pre-trained with MAE method then fine-tuned on the specific task.
A lightweight ViT model after generic distillation is set as the student $f^s$.
As concrete the loss function is depend on specific tasks, we denote $\gL_{\text{Task}}$ as the task loss function, $\gL_{\text{KD}}$ as the task-specific distillation loss function.
Combining the task loss with task-specific distillation loss, we have a joint loss to optimize the student model, as
\begin{align}
\gL_{\text{SD}} = \gL_{\text{Task}}(f^s(\vx),Y) + \beta\gL_{\text{KD}}(f^s(\vx),f^t(\vx)),
\label{eq:sd_objective}
\end{align}
where $Y$ is the ground truth and $\beta$ is the regularization factor (Refer to Appendix~\ref{hyperparameters} for details).

\subsection{Analysis}
The proposed two-stage approach is more plausible than commonly used single-stage methods, which can be justified from the perspective of mutual information~\cite{VariationalID}.
The knowledge distillation can be generally interpreted as a procedure to maximize the mutual information $\mathcal{I}$ of a teacher model ($f^t$) and a student model ($f^s$).
Denote the parameters of the student model as $\theta^s$, the pre-training dataset as $X$ and the fine-tuning dataset as $\hat{X}$.
The single-stage task-specified distillation is interpreted as
\begin{equation}
\argmax_{\theta^s} \mathcal{I}_{\theta^s,\theta^t}(f^t,f^s|\hat{X}),
\label{eq:mi_single}
\end{equation}
which maximizes the mutual information between the teacher model $f^t$ and the student model $f^s$ conditional on the fine-tuning dataset $\hat{X}$.
The proposed G2SD is interpreted as
\begin{equation}
\begin{split}
\argmax_{\theta^s} &\ \mathcal{I}_{\theta^s,\theta^t}(f^t,f^s|X)+\mathcal{I}_{\theta^s,\theta^t}(f^t,f^s|\hat{X})\\
&-\mathcal{I}_{\theta^s,\theta^t}(f^t,f^s|(X,\hat{X})),
\label{eq:mi_g2sd}
\end{split}
\end{equation}
which maximizes the mutual information between the teacher model $f^t$ and the student model $f^s$ conditional on both the pre-training data $X$ and the fine-tuning dataset $\hat{X}$.
Obviously, the mutual information defined by~\cref{eq:mi_g2sd} is larger than that by~\cref{eq:mi_single}, which implies more information can be transferred by our G2SD approach.

\section{Experiments}
\label{sec:exp}

\subsection{Setting}

\paragraph{Datasets.} 

The generic distillation is conducted on ImageNet-1k~\cite{imagenet} training set with 1.2M images. 
Following self-supervised recipes~\cite{mae}, we do not use the label information, so that lightweight models focus on soaking up the task-agnostic representations.
In specific distillation, the models are fine-tuned from the previous stage on ImageNet-1k~\cite{imagenet}, MS COCO~\cite{mscoco} and ADE20K~\cite{ade20k} datasets.
%

\begin{table}[t]
\caption{Top-1 accuracy on ImageNet-1k.}
\centering
\resizebox{0.95\linewidth}{!}{
\begin{tabular}{lccc}
\toprule
\bf Method & \bf Teacher & \bf \#Param(M)   & \bf Acc (\%) \\ \midrule
DeiT-Ti~\cite{deit} & \multirow{9}{*}{N/A} & 5 & 72.2 \\ 
MobileNet-v3~\cite{MobileNetV3} & & 5  & 75.2 \\
ResNet-18~\cite{resnet} &  & 12  & 69.8 \\
DeiT-S~\cite{deit} &  & 22     &  79.8 \\
BEiT-S~\cite{beit} & & 22   & 81.7 \\
CAE-S~\cite{cae} & & 22   & 82.0 \\ 
DINO-S~\cite{dino} & & 22 &  82.0 \\
iBOT-S~\cite{ibot} & & 22 &  82.3 \\
ResNet-50~\cite{resnet} & & 25  & 76.2 \\
Swin-T~\cite{swin} & & 28 & 81.3 \\
ConvNeXt-T~\cite{convnext} & & 29  & 82.1 \\
\midrule
DeiT-Ti\alambic~\cite{deit} &  & 6  & 74.5 \\ 
DeiT-S\alambic~\cite{deit} & RegNetY- & 22 & 81.2 \\
DearKD-Ti~\cite{Chen2022DearKDDE} & 16GF & 6  & 74.8 \\ 
DearKD-S~\cite{Chen2022DearKDDE} &  & 22   & 81.5 \\ 
\midrule
Manifold-Ti~\cite{manifold_distill} &  & 6 & 75.1 \\
Manifold-S~\cite{manifold_distill} &  CaiT- & 22 & 81.5 \\
MKD-Ti~\cite{MetaKD} & S24 & 6 & 76.4 \\
MKD-S~\cite{MetaKD} &  & 22 & 82.1 \\
\midrule
SSTA-Ti~\cite{SSTA} & DeiT-S & 6  & 75.2 \\
SSTA-S~\cite{SSTA} & DeiT-B & 22  & 81.4 \\ 
\midrule
DMAE-Ti~\cite{DMAE} & \multirow{4}{*}{MAE-B}  & 6   & 70.0 \\
DMAE-S~\cite{DMAE} & & 22  & 79.3 \\
\our{-Ti (\textit{ours})} & & 6    &  77.0 \\
\our{-S (\textit{ours})} & & 22    & \bf 82.5 \\ 
\bottomrule
\end{tabular}
}
\label{tab:mainresult-classification}
\end{table}  

\paragraph{Implementation details.} 
In generic distillation stage, the MAE pre-trained ViT-Base model~\cite{mae} is employed as the teacher. The student model is trained for 300 epochs using the AdamW optimizer~\cite{adamw}, learning rate 2.4e-3, weight decay 0.05, batch size 4096, and image resolution 224×224. Unless specified, the mask ratio is set to 75\% and the student decoder contains 4 Transformer blocks with 128 and 256 dimensions for ViT-Tiny and ViT-Small, respectively. 

In task-specific distillation stage, the student decoder is discarded while the encoder is utilized as backbone to extract feature for various tasks, as do in MAE~\cite{mae}. We use the official or re-implemented MAE fine-tuned model as the teacher. To avoid deteriorating the general representations obtained from the previous stage, a layer decay schedule is adopted to train the student model for all downstream tasks.

For image classification, we take a fine-tuned ViT-base model as the teacher, which is officially released by MAE~\cite{mae} and achieves 83.6\% top-1 accuracy. Following DeiT~\cite{deit} distillation recipe, we append a distillation token to the student model for token-based distillation and use the hard decision of the teacher as the distillation label. The student model is trained for 200 epochs.

\begin{table}[t]
\caption{Object detection and instance segmentation results on the MS COCO dataset.}
\centering
\resizebox{0.9\linewidth}{!}{
\begin{tabular}{@{}lccc}
\toprule
\bf Method & \bf \#Param(M)  & \bf AP$^{bbox}$ &\bf AP$^{mask}$\\ \midrule
\multicolumn{4}{l}{\textit{Mask R-CNN~\cite{maskrcnn}, 36 epochs + Multi-Scale}} \\
CAE-S~\cite{cae} & 46.1  & 44.1 & 39.2 \\
ViT-Adapter-T~\cite{ViTadapter} & 28.1  & 46.0 & 41.0 \\ 
Swin-T~\cite{swin} & 47.8  & 46.0 & 41.6 \\ 
ConvNeXt-T~\cite{convnext} & 48.1  & 46.2 & 41.7 \\
imTED-S~\cite{imted} & 30.1 & 48.0   & 42.8\\ 
ViT-Adapter-S~\cite{ViTadapter} & 47.8  & 48.2 & 42.8 \\
\midrule
\multicolumn{4}{l}{\textit{ViTDet~\cite{vitdet}, 100 epochs + Single-Scale}} \\
DeiT-S\alambic ~\cite{deit}  & 44.5 & 47.2 & 41.9 \\
DINO-S ~\cite{dino} & 44.5 & 49.1 & 43.3 \\
iBOT-S ~\cite{ibot} & 44.5 & 49.7 & 44.0 \\
\our{-Ti (\textit{ours})} & 27.7   &  46.3 &  41.6 \\
\our{-S (\textit{ours})} & 44.5 & \bf 50.6 & \bf 44.8 \\ 
\bottomrule
\end{tabular}
}
\label{tab:mainresult-detection}
\end{table}

\begin{table}[t]
\caption{ADE20K validation results using UperNet~\cite{upernet}. The input image resolution is 512$\times$512.}
\vspace{-0.5em}
\centering
\resizebox{0.75\linewidth}{!}{
\begin{tabular}{lcc}
\toprule
\bf Method & \bf \#Param(M)  &\bf mIoU \\ 
\midrule
ViT-Adapter-Ti~\cite{ViTadapter} & 36.1  &  42.6  \\
Swin-T~\cite{swin} & 59.9   &  44.5  \\ 
ConvNeXt-T~\cite{convnext} & 60   &  46.0   \\ 
ViT-Adapter-S~\cite{ViTadapter} & 57.6  &  46.6  \\
\midrule
DINO-S ~\cite{dino} & 42.0 & 44.0 \\
iBOT-S ~\cite{ibot} & 42.0 & 45.4 \\
\our{-Ti (\textit{ours})} & 11.0   &  44.5 \\
\our{-S (\textit{ours})} & 42.0   & \bf 48.0  \\ 
\bottomrule
\end{tabular}
}
\label{tab:mainresult-segmentation}
\vspace{-1em}
\end{table}

\begin{table*}[ht]
\caption{Ablation study on single-stage and two-stage distillation methods,
where \ourgd{} denotes \textbf{only} performing generic distillation (\ie, without specific distillation) and MAE\scalebox{0.9}{\alambictwocol} means performing task-specific distillation during fine-tuning phase of MAE~\cite{mae}.}
\vspace{-0.5em}
\centering
\small
\scalebox{0.9}{
\begin{tabular}{@{}lcccccccc}
\toprule
\multirow{2}{*}{\bf Method} & \bf Params & \bf Throughout & \bf Generic & \bf Specific  & \bf ImageNet-1k &  \multicolumn{2}{c}{\bf MS COCO} & \bf ADE20k   \\
& \bf(M) & \bf(Images/s) & \bf Distillation & \bf Distillation  & \bf Top-1 Acc (\%) & \bf AP$^{bbox}$ & \bf AP$^{mask}$ & \bf mIoU  \\
\midrule
\textit{\gc{Teacher: ViT-Base}} & \gc{86.57} & \gc{1.0$\times$} & \gc{N/A} & \gc{N/A} & \gc{83.6} & \gc{51.6} & \gc{45.9} & \gc{48.3}  \\
\midrule
\multicolumn{9}{l}{\textit{Student: ViT-Tiny}} \\
MAE~\cite{mae} & 5.72 & 5.84$\times$ & \xmark & \xmark & 75.2 & 37.9 & 34.9 &  36.9 \\ 
MAE\alambictwocol~\cite{mae}  & 5.91 & 5.74$\times$ & \xmark & \cmark & 75.9  &43.5 & 39.0& 42.0  \\
\ourgd{} (\textit{ours})  & 5.72 & 5.84$\times$ & \cmark & \xmark & 76.3 & 44.0 & 39.6 & 41.4 \\ 
\our{} (\textit{ours})  & 5.91 & 5.74$\times$ & \cmark & \cmark & \bf 77.0 & \bf 46.3 & \bf 41.3 & \bf44.5 \\ 
\midrule
\multicolumn{9}{l}{\textit{Student: ViT-Small}} \\
MAE~\cite{mae}  & 22.05 & 2.62$\times$ & \xmark & \xmark & 81.5 & 45.3 & 40.8 &  41.1 \\ 
MAE\alambictwocol~\cite{mae}  & 22.44 & 2.58$\times$ & \xmark & \cmark & 81.9 & 48.9 & 43.5 & 44.9 \\
\ourgd{} (\textit{ours}) & 22.05 & 2.62$\times$  & \cmark & \xmark & 82.0 & 49.9 & 44.5 & 46.2 \\ 
\our{} (\textit{ours}) & 22.44 &  2.58$\times$ & \cmark & \cmark & \bf 82.5 & \bf 50.6 & \bf 44.8 & \bf 48.0 \\ 
\bottomrule
\end{tabular}
}
\label{tab:abla:two_stage}
\end{table*}

For object detection and instance segmentation tasks, we follow the ViTDet~\cite{vitdet} framework, where the official ViTDet-Base~\cite{vitdet} model are used as the teacher.
The Feature-Richness Score method~\cite{frs} is adopted to stress important features that are distilled from the teacher to the student model. Student models are trained with batch size 64 for 100 epochs. The input image resolution is $1024\times1024$.

For semantic segmentation, we use UperNet~\cite{upernet} task layers and distill the model for 160K iterations. Due to the absence of officially released model weights, we fine-tune the MAE pre-trained ViT-Base model on ADE20k by using the BEiT~\cite{beit} semantic segmentation codebase to get teacher model, which achieves 48.3 mIoU, is comparable to MAE official report. During specific distillation, besides the supervision from the ground-truth, activation maps from the student and the teacher are aligned $w.r.t.$ the channel dimension~\cite{cwd}.

\subsection{Main Results}
\paragraph{Image Classification.}
In ~\cref{tab:mainresult-classification}, \our{} is compared with
1) supervised methods including  MobileNet-v3~\cite{MobileNetV3}, ResNet~\cite{resnet,Wightman2021ResNetSB}, DeiT~\cite{deit,deit3}, Swin Trasnformer~\cite{swin} and ConvNeXt~\cite{convnext};
2) self-supervised methods upon ViT-Small, like BEiT~\cite{beit} and CAE~\cite{cae}; and 3) distillation methods upon vanilla ViTs, like DeiT\alambic~\cite{deit}, DearKD~\cite{Chen2022DearKDDE}, Manifold~\cite{manifold_distill}, MKD~\cite{MetaKD}, SSTA~\cite{SSTA} and DMAE~\cite{DMAE}.
\our{} achieves 82.5\% top-1 accuracy, which outperforms CNN-based ConvNeXt by 0.4\%, by using fewer parameters (22M \vs 29M).
\our{} consistently outperforms self-supervised methods, BEiT and CAE, by 0.8\% and 0.5\%, respectively.
Compared with those distillation methods, \our{} shows the superiority.
Remarkably, with the limited parameters ($\sim$6M), \our{} reports a substantial gain compared to DeiT-Ti\alambic and carefully designed MobileNet-v3.

\paragraph{Object Detection and Instance Segmentation.} 
In~\cref{tab:mainresult-detection}, we report AP$^{bbox}$ for object detection  and AP$^{mask}$ for instance segmentation. We compare \our{} with some popular methods on various backbone network: 1) vanilla ViT, like CAE~\cite{cae}, ViT-Adapter~\cite{ViTadapter}, imTED~\cite{imted}; 2) elaborately designed architecture, like CNN based ConvNeXt~\cite{convnext} and hierarchical designed Swin Transformer~\cite{swin}. One can see that \our{}-S, with fewer parameters, obtains more than 4.4 AP$^{bbox}$ gains compared with ConvNeXt-T and Swin-T, which contain many inductive bias. Compared with CAE-S, which benefits from masked image modeling, \our{} also show the extraordinary superiority. Moreover, \our{}-S significantly outperforms imTED-S by 2.6 AP$^{bbox}$ on object detection and 2 AP$^{mask}$ on instance segmentation, where imTED-S uses pre-trained MAE encoder as backbone and pre-trained MAE decoder as task layers.

\paragraph{Semantic Segmentation.}
In~\cref{tab:mainresult-segmentation}, \our{} is compared with ViT-Adapter~\cite{ViTadapter}, ConvNeXt~\cite{convnext} and Swin Transformer~\cite{swin}.
\our{-S} outperforms all the compared methods by at least 1.4 mIoU, where ViT-Adapter elaborately modifies the model architecture for adapting dense prediction tasks. Remarkably, only using 11M parameters, \our{-Ti} achieves 44.5 mIoU, which pushes the performance of lightweight ViT models to a new height.

\begin{table}[t]
\caption{Ablation study on generic distillation targets.  $\ve^{t}_{i}$, $\hat\vz^{t}_{i}$ and $\gR(\ve^{t}_{i})$ respectively denote teacher encoder features, teacher decoder features, the relation among teacher encoder features. \#5 is the default setting.
}
\vspace{-0.5em}
\centering 
\footnotesize
\scalebox{0.97}{
\begin{tabular}{l|cccc|cc}
\toprule
\multirow{2}{*}{\textit{Target}} & $\ve^{t}_{i}$ &  $\gR(\ve^{t}_{i})$ & $\hat\vz^{t}_{i}$ & $\hat\vz^{t}_{i}$ & Accuracy & mIoU \\ 
& \scriptsize $i\in \gV$ & \scriptsize $i\in \gV$ & \scriptsize $i\in \gV$ & \scriptsize $i\in \gM$ & (\%) & (\%) \\ 
\midrule
\textit{\#1} &   \cmark   &              &             &              &  81.60  &  43.69  \\
\textit{\#2} &            &    \cmark    &             &              & 81.45   &  43.64   \\
\textit{\#3} &            &              &             &    \cmark    & 81.96   & 45.20\\ 
\textit{\#4} & \cmark     &              &             &        \cmark      &   81.85 & 44.12 \\ 
\textit{\#5} &            &              &  \cmark     &     \cmark   &  \bf 81.99  & \bf 46.19\\  
\bottomrule
\end{tabular}
}
\label{tab:abla:general_target}
\vspace{-1em}
\end{table}

\subsection{Ablation Studies: Single-stage \textbf{\vs} Two-stage}
In~\cref{tab:abla:two_stage}, comprehensive experiments are conducted to compare single-stage and two-stage distillation methods. The teacher models include: 1) pre-trained MAE ViT-Base model for generic distillation; 2) fine-tuned MAE models on ImageNet-1k, MS COCO and ADE20k for specific distillation, which respectively reach 83.6\% top-1 accuracy, 51.6 AP$^{bbox}$, 45.9 AP$^{mask}$ and 48.3 mIoU.
The student models are vanilla ViT-Tiny and ViT-Small, which are initialized from self-supervised method MAE~\cite{mae} and \our{}.
We denote MAE\alambic as model pre-trained with MAE and fine-tuned with task specific distillation. 
MAE ViT-Small model is pre-trained for 300 epochs, by using the official codebase.

When specific distillation is not used, \ourgd{} outperforms MAE by a large margin, \eg, 49.9 \vs 45.3 AP$^{bbox}$ and 46.2 \vs 41.1 mIoU, which benefits from the transferred task-agnostic knowledge. After activating specific distillation, both MAE\alambic and \our{} boost their performances, \eg, \our{-S} achieves 0.7 AP$^{bbox}$ gains and 1.8 mIoU gains, which are attributed to discriminative representation configuration.
In conclusion, \our{} outperforms MAE and MAE\alambic across model sizes and datasets, validating the superiority of our two-stage distillation approach.

\subsection{Ablation Studies: Generic Distillation}
\paragraph{Target Configuration.}
We investigate the impact of target feature selection in generic distillation stage and report the results in \cref{tab:abla:general_target}. All models are trained under the same recipe and evaluated on ImageNet-1k and ADE20k.
From~\cref{tab:abla:general_target} (\#5), one can see that aligning student decoder features with teacher decoder's hidden features at both visible and masked patches achieves the best results, \eg, 81.99\% top-1 accuracy on ImageNet-1k and 46.19 mIoU on ADE20k.

Transferring from teacher encoder to student encoder is the most straightforward method, as shown in~\cref{tab:abla:general_target} (\#1), but it only reaches 43.69 mIoU on ADE20k. The reason lies on that it overlooks the context understanding ability, which is beneficial for dense prediction tasks.
Distilling the relation among tokens is popular and effective in NLP~\cite{MiniLMv2}.
We thus conduct experiments using the self-attention relation of teacher encoder as distillation target, and find that the student only obtains 81.45\% top-1 accuracy on ImageNet-1k, ~\cref{tab:abla:general_target} (\#2).

In~\cref{tab:abla:general_target} (\#3), we align the student decoder features with those of the teacher decoder on the masked positions.
In this way, the student respectively gets 0.36\% accuracy and 1.51 mIoU gains on ImageNet and ADE20k compared to~\cref{tab:abla:general_target} (\#1), which verifies the superiority of learning the context understanding capacity.
Furthermore, we simultaneously calculate alignment loss on encoder features at visible patches and on decoder features at masked patches in~\cref{tab:abla:general_target} (\#4), which is a more direct approach to let student inherit the feature extracting and context understanding capability of teacher, compared with ~\cref{tab:abla:general_target} (\#5).
Unfortunately, the student performs worse than only calculating alignment loss on decoder features at masked patches, \eg, 44.12 \vs 46.19 mIoU on ADE20k.

\begin{table}[t]
\caption{Ablation on the mask ratio (\textit{top}) and  target layer of the teacher model used for distillation (\textit{bottom}).}
\centering
\small
\begin{tabular}{l|ccccc}
\toprule
Mask ratio & 0.05 & 0.25  & 0.55 & 0.75 & 0.9  \\ \midrule
Top-1 Acc(\%) & 81.7 &	81.7 &	81.6 &	\bf 82.0 & 	81.8 \\ \midrule \midrule
Layer Index & 1 & 2  & 4 & 6 & 8 \\ \midrule
Top-1 Acc(\%) & 81.6 &	81.8 & \bf	82.0 &	81.8 & 	81.7 \\
\bottomrule
\end{tabular}
\label{tab:abla_mask}
\end{table}

\paragraph{Mask Ratio.}
A high mask ratio (75\%) works well in MAE~\cite{mae}, but the suitable mask ratio in generic distillation still needs to be explored. In general, predicting masked features is more challenging than predicting pixels. However, the observations are consistent with the teacher MAE, as illustrated in~\cref{tab:abla_mask} (\textit{top}), where a high mask ratio tends to generate good results.
The reason may be that the teacher model can express itself to the greatest extent when the mask ratio is consistent with the MAE pre-training phase.

\paragraph{Target Layer.}
A sufficiently deep decoder is essential for the fine-tuning performance in MAE~\cite{mae}. We study the impact of which decoder layer is the best target layer~\cref{tab:abla_mask} (\textit{bottom}).
One can see that using features of the 4-th layer as distillation targets for \our{} yields better accuracy. This can be explained that the last several layers in decoder are more specialized for low-level information (\eg, pixel values) reconstruction while the first several layers in a decoder can't produce enough general representations.

\paragraph{Decoder Design.} 
We study how the performance varies with decoder depth and width, where depth and width respectively denote the number of Transformer blocks and the embedding dimension of each Transformer block. 
As demonstrated in \cref{tab:abla:decoder_design}, the student decoder of width 256 and depth 4 yields optimal results, in terms of image classification. When the student decoder is heavy, the student encoder can be ``lazy" to pursue good features as the decoder is competent for both feature extraction and image reconstruction.

\begin{table}[t]
\caption{Ablation study on the width and depth (D) of the student decoder. The depth and width of the teacher's decoder are 8 and 512, respectively.
}
\centering
\small
\begin{tabular}{c|cc|cc|cc}
\toprule
Width & D & Acc(\%) & D & Acc(\%) & D & Acc(\%) \\ 
\midrule
128 & \multirow{3}{*}{2} & 81.9 & \multirow{3}{*}{4} & 81.8 & \multirow{3}{*}{8} & 81.7\\
256 &  & 81.7 &  & \bf 82.0 &  & 81.7 \\
512 &  & 81.8 &  & 81.7 &  & 80.3 \\
\bottomrule
\end{tabular}
\label{tab:abla:decoder_design}
\end{table}

\begin{figure}[t]
\centering
\includegraphics[width=1.0\linewidth]{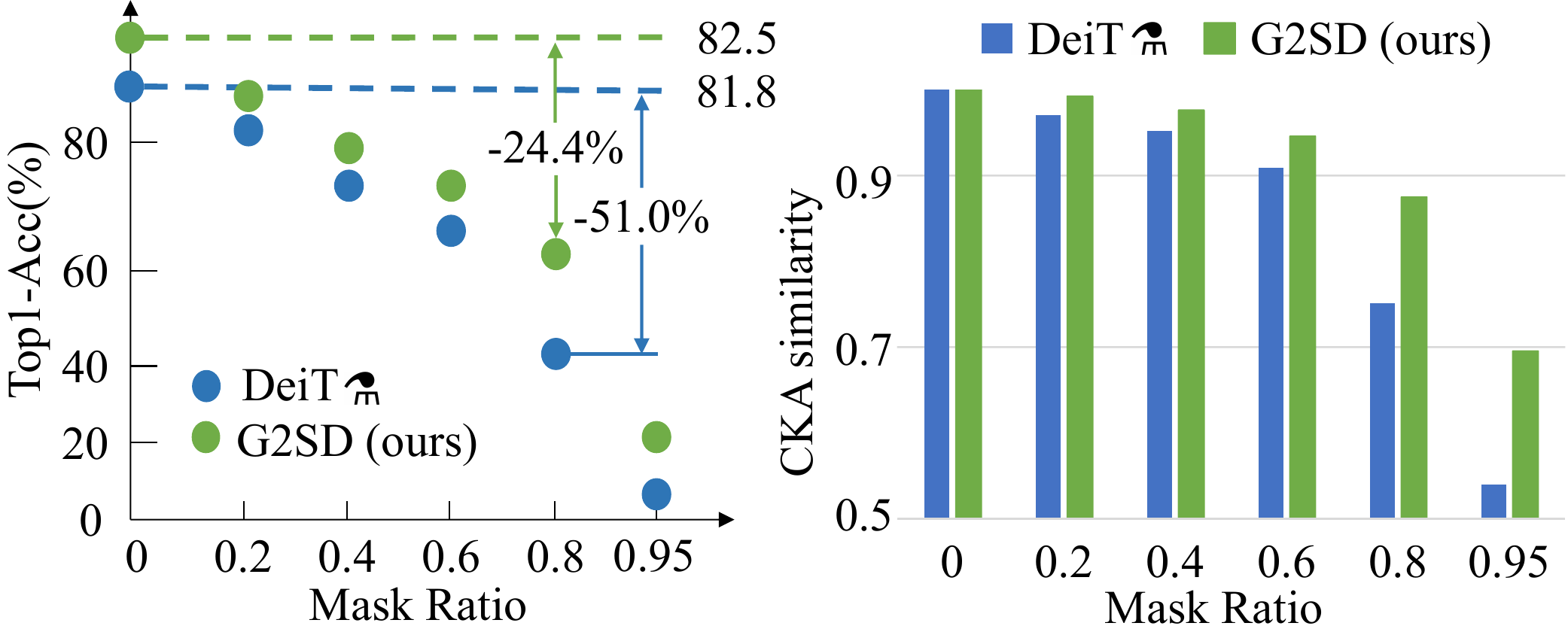}
\caption{Performance degradation (\textit{left}) and CKA similarity (\textit{right}) between the representations generated by the complete image and the corrupted image with various mask ratios.}
\label{fig:occlusion}
\end{figure}
\begin{table}[t]
\caption{Robustness evaluation. ``IN" is short for ImageNet.}
\centering
\resizebox{1.0\linewidth}{!}{
\begin{tabular}{@{}lccccc}
\toprule
\bf Methods & \bf IN & \bf IN-A  & \bf IN-R & \bf IN-S & \bf IN-V2  \\ \midrule
\textit{\gc{Teacher: ViT-Base}} &  \gc{83.6} & \gc{35.9} & \gc{48.3} & \gc{34.5} & \gc{73.2} \\
\midrule
\multicolumn{5}{l}{\textit{Student: ViT-Tiny}} \\
DeiT\alambic~\cite{deit} & 75.3 & 9.5 & 36.2 & 23.4 & 63.3\\
MAE\alambic~\cite{mae}   & 75.9 & 10.9  & 38.7 &  \bf 26.3  & 64.7\\ 
\our{} (ours)  & \bf 77.0  &  \bf12.9 &  \bf39.0 & 25.9& \bf65.6\\ 
\midrule
\multicolumn{5}{l}{\textit{Student: ViT-Small}} \\
DeiT\alambic~\cite{deit} & 81.8 & 24.2 & 45.9 & 32.1 &71.1\\
MAE\alambic~\cite{mae}   & 81.9 & 26.6 & \bf 46.8 & \bf 34.3 & 71.1\\ 
\our{} (ours)    & \bf 82.5 &  \bf 29.4 &  \bf 46.8 & 33.6 & \bf72.1\\ 
\bottomrule
\end{tabular}
}
\label{tab:Robust}
\end{table}

\subsection{Analysis}
\our{} is compared with MAE\alambic and DeiT\alambic in terms of occlusion invariance, representation similarity and robustness, which indicate that it learns representations general to downstream tasks. DeiT\alambic denotes performing task-specific distillation by replacing the original teacher with the fine-tuned MAE-Base model.  For fair comparison, we set the total training epochs of the three methods to be same (500 epochs). The major difference between those three methods is initialization, $i.e.$, \our{} is initialized from generic distillation, MAE\alambic from MAE pre-training, and DeiT\alambic from scratch.

Centered Kernel Alignment~\cite{cka} (CKA) is a preferred metric evaluating normalized similarity between two feature maps or representations, and it is invariant to the orthogonal transformation of representations and isotropic scaling.
We calculate CKA scores to analyze the occlusion invariance and representation similarity in the following.

\paragraph{Occlusion Invariance.} 
\label{sec:sec:occlusion}

Masked autoencoders are verified to learn occlusion invariant features in~\cite{UnderstandMAE2022}. In~\cref{fig:occlusion} (\textit{left}), we directly evaluate the performance of DeiT\alambic and \our{} under various mask ratios on the ImageNet-1k validation set. \our{} decreases about 24\% while DeiT\alambic decreases about 51\% when the mask ratio is 80\%. 
In~\cref{fig:occlusion} (\textit{right}), we calculate the CKA similarity between masked image representations and complete image representations, and find that \our{} can obtain higher CKA scores than DeiT\alambic. These observations suggest that \our{} preserves more occlusion invariance than the single-stage method (\eg, DeiT\alambic).

\begin{figure}[t]
\centering
\includegraphics[width=1.0\linewidth, height=0.85\linewidth]{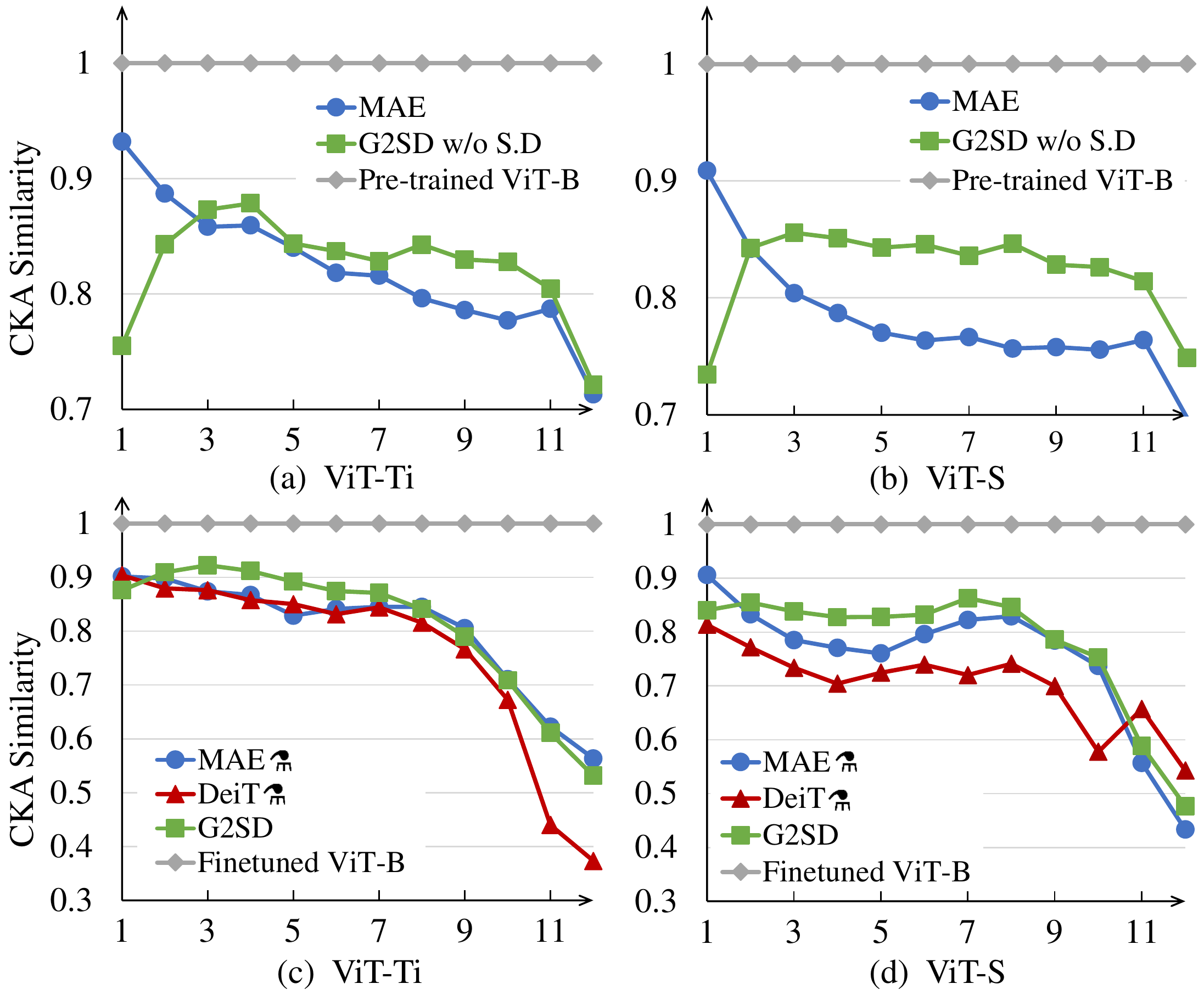}
\caption{
CKA similarity between representations generated by pre-trained MAE-B with: (a) pre-trained MAE-Ti and \our{-Ti} w/o S.D, and (b)  MAE-S and \our{-S} w/o S.D. CKA similarity between representations generated by fine-tuned MAE-B with: (c) fine-tuned MAE-Ti\alambic, DeiT-Ti\alambic, and \our{-Ti}, and (d) fine-tuned MAE-S\alambic, DeiT-S\alambic, and \our{-S}.
\textit{x-axis} denotes network depth.}
\label{fig:CKA}
\end{figure}

\paragraph{Representation Similarity.} 
In~\cref{fig:CKA} (a) and (b), representations generated
by \ourgd{} is more similar with pre-trained MAE-B than pre-trained MAE-S, indicating that generic distillation enables better features than simply reconstructing pixels.
Furthermore, after task-specific distillation, \our{} consistently obtains higher CKA scores than MAE\alambic and DeiT\alambic, as illustrated in~\cref{fig:CKA} (c) and (d), implying that generic distillation provides a favored initialization for specific distillation.

\paragraph{Robustness.} 
This is evaluated by testing the trained classifier on several ImageNet variants including ImageNet-A~\cite{adversarial2021}, ImageNet-R~\cite{rendition2021}, ImageNet-S~\cite{sketch2019} and ImageNet-V2~\cite{imagenetv2}. From~\cref{tab:Robust}, one can see that \our{} outperforms the compared methods, which implies better generalization capability. In other words, the proposed G2SD encourages the small student model to maintain the generalization capability of teacher model endowed by the generic self-supervised method, as much as possible.

\section{Conclusion}

We proposed a two-stage distillation approach, termed generic-to-specific distillation (G2SD), to tap the potential of lightweight ViTs under the supervision of pre-trained large models.
For generic distillation, we further designed a simple-yet-effective distillation strategy by aligning students' predictions with latent features of large masked autoencoders at both masked and visible patches.
With two-stage distillation, the task-agnostic and task-specific knowledge of large models were transferred to lightweight ones.
Extensive experiments on image classification, object detection, and semantic segmentation validated the performance of the proposed G2SD approach, with striking contrast with state-of-the-art methods.
This study has built a solid baseline for the two-stage vision model distillation.

\noindent\textbf{Acknowledgement.}
This work was supported by National Natural Science Foundation of China (NSFC) under Grant 62225208, 62171431 and 61836012, and the Strategic Priority Research Program of Chinese Academy of Sciences under Grant No. XDA27000000. 


{\small
\bibliographystyle{ieee_fullname}
\bibliography{main}
}

\appendix

\section{Hyperparameters}\label{hyperparameters}

\subsection{Image Classification}
For distillation, as in~\cite{deit}, we added a learnable distillation token, which is combined with the cls token to produce final predictions in the inference phase. In experiments, the data augmentation and optimizer follow the fine-tuning recipe of MAE \cite{mae}, while the learning rate, training epochs and layer-wise learning-rate decay are specified. 
For models training from scratch (e.g., DeiT\alambic{}), we set the layer decay value as 1.0, which means no layer decay is adopted. 
For pre-trained models (e.g., MAE~\cite{mae}, G2SD),  we set the layer decay value to 0.75 and training epochs to 200.

\begin{table}[h!]
\caption{
Hyperparameters for distilling on ImageNet-1K.
}
\label{tbl:pretrain:hyperparams}
\centering
\small
\scalebox{0.98}{
\begin{tabular}{l|cc}
\toprule
\bf \multirow{2}{*}{Hyperparameters} & \bf Value & \bf Value \\
            & \bf(Fine-tuning) & \bf (From scratch) \\
\midrule
Training epochs & 200 & 500 \\
Base learning rate & 1e-3 & 2.5e-4\\
Layer decay & 0.75 & 1.0 \\
\midrule
Warm up epochs & \multicolumn{2}{c}{5}\\
Label smoothing & \multicolumn{2}{c}{0.1} \\
Mixup & \multicolumn{2}{c}{0.8} \\
Cutmix & \multicolumn{2}{c}{1.0} \\
Drop path & \multicolumn{2}{c}{0.0} \\
Batch size & \multicolumn{2}{c}{1024} \\
Weight decay & \multicolumn{2}{c}{0.05} \\
Optimizer & \multicolumn{2}{c}{AdamW } \\
Learning rate schedule & \multicolumn{2}{c}{Cosine decay} \\
Augmentation & \multicolumn{2}{c}{RandAug(0,0.5)} \\
Optimizer momentum & \multicolumn{2}{c}{$\beta_1$, $\beta_2$ = 0.9, 0.999 } \\
\bottomrule
\end{tabular}
 }
\end{table}

\subsection{Object Detection and Instance Segmentation}

In the experiments, we adopt the official codebase\footnote{\url{https://github.com/facebookresearch/detectron2/tree/main/projects/ViTDet}} and follow the settings used in ViTDe ~\cite{vitdet}. The total batch size is set to 64 (8 images per GPU).
The learning rate is set to $1e^{-4}$, the backbone's drop path rate is $0.1$, and the distill warm step is 500. The overall training target is the same as ~\cite{frs}: $L = L_{GT} +\alpha L_{FPN} + \beta L_{head}$, where $\alpha$ and $\beta$ are respectivvely set to $0.001$ and $0.1$.                                             
\subsection{Semantic Segmentation}
In this experiment, we adopt the BEiT's segmentation codebase\footnote{\url{https://github.com/microsoft/unilm/beit}} and set the total batch size to 32 (4 images per GPU). The backbone's drop path rate is $0.1$. The layer decay rate is 0.75. The learning rate of ViT-Small and ViT-Tiny are respectively set to $2e^{-4}$ and $5e^{-4}$. We set the temperature parameter $\tau = 1$, the loss weight $\alpha = 3$ for the logits map distillation.

\section{Training Time and Efficiency}
As shown in Table~\ref{tab:longer_schedule}, G2SD outperforms DeiT~\cite{deit} and DeiT\alambic~\cite{deit}, which have a longer training schedule (500 epochs). The teacher of DeiT\alambic is the same as \our{'s}. In the generic distillation stage, since the input of \our{} is a masked image (75\% patches are discarded), the training time per epoch is less than DeiT (which computes the whole image).

\begin{table}[h]
\caption{\our{} $vs$ DeiT. The total training epochs is 500.}
\label{tab:longer_schedule}
\centering
\small
\resizebox{\linewidth}{!}{
\begin{tabular}{c|c|c|c|c}
\toprule
\bf Methods & \bf 1-st stage & \bf 2-nd stage & \bf Time & \bf Top-1 Acc (\%) \\ \midrule
\our{} &  G.D 300 epochs & S.D 200 epochs & 71 h & 82.5 \\ \midrule
DeiT\alambic & \multicolumn{2}{c|}{Supervised+Distillation 500 epochs} & 112 h & 81.7 (\textbf{-0.8}) \\
\midrule  
DeiT & \multicolumn{2}{c|}{Supervised 500 epochs} & 53 h&  81.4 (\textbf{-1.1}) \\ 
\bottomrule
\end{tabular}
}
\end{table}

\section{Detection Performance with ViTDet}

For the lack of official Mask-RCNN~\cite{maskrcnn} results and checkpoints of MAE~\cite{mae}, we choose ViTDet~\cite{vitdet} as the detector. 
In Table~\ref{tab:vitdet}, the backbone models are initialized from various supervisions, \eg, supervised methods (DeiT~\cite{deit}), distilled methods (DeiT\alambic~\cite{deit} and \our{}) and self-supervised methods (DINO~\cite{dino} and iBoT~\cite{ibot}).
From Table~\ref{tab:vitdet}, \our{} significantly outperforms competitors on performance and convergence speed.


\begin{table}[h]
\caption{Performance on MS COCO using the ViTDet framework~\cite{vitdet}, which is trained for 100 epochs with single-scale input (1024$\times$1024).}
\label{tab:vitdet}
\centering
\small
\resizebox{\linewidth}{!}{
\begin{tabular}{l|c|cc}
\toprule
\bf Methods (Supervision) & \bf ImageNet Acc (\%) & \bf AP$^{bbox}$ & \bf AP$^{mask}$ \\ \midrule
DeiT-S (sup., 300e) & 79.9 & 45.7 & 40.7 \\
DeiT-S\alambic (sup.\&distill., 300e) & 81.2 & 47.2 & 41.9 \\
DeiT-S (sup., 500e) & 81.4 & 46.9 & 41.6\\
DINO-S (self-sup., 3200e) & 82.0 & 49.1 & 43.3 \\
iBOT-S (self-sup., 3200e) & 82.3 & 49.7 & 44.0 \\ \midrule
G2SD-S (w/o S.D, 300e) & 82.0 & 49.9 & 44.5 \\
G2SD-S (300e) & 82.5 & 50.6 & 44.8
 \\ \bottomrule
\end{tabular}
} 
\end{table}

\section{More Ablations on Target Configuration}

In Table~\ref{tab:abla:general_target}, we have conducted ablation studies on intermediate features as generic distillation targets.
Compared with using intermediate features as distillation targets, taking the teacher's prediction as distillation objective~\cite{hinton2015distilling,deit} is also a popular alternative. 
Therefore, we take the MAE's predictions as the generic distillation targets in Table~\ref{tab:abla:more_general_target}. When taking the MAE's predictions as the targets for masked positions, the performance drops to 81.4\% (without specific distillation) and 81.8\% (with specific distillation).
This observation is consist with the results in Table~\ref{tab:abla_mask} (\textit{bottom}), where the last several layers in decoder are more specialized for low-level information reconstruction task.

\begin{table}[t]
\caption{Ablation study of distillation targets on ImageNet-1k. `S.D' is short for specific distillation.
}
\label{tab:abla:more_general_target}
\centering
\small
\scalebox{0.8}{
\begin{tabular}{c|c|c}
\toprule
\bf Distillation targets & \bf W/O S.D Acc (\%) & \bf W S.D Acc (\%)\\ \midrule
Our default settings & \bf 82.0 & \bf 82.5 \\ \midrule
MAE's reconstructions & 81.4 & 81.8\\ \midrule
MAE's reconstructions + GT & 81.5 & 81.7
 \\ \bottomrule
\end{tabular}
}
\vspace{-0.5em}
\end{table}



\end{document}